\documentclass{article}
\usepackage{spconf,amsmath,graphicx}
\usepackage[hypertexnames=false]{hyperref}
\usepackage{float}
\usepackage{booktabs,multirow}
\usepackage[table]{xcolor}

\title{C3M: Cross-Session Multimodal Memory Maintenance \\ for Long-Horizon Tasks}
\name{%
\begin{tabular}{c}
Xueshu Chen\textsuperscript{1*}, Yan Wang\textsuperscript{2*}, Zihao Xue\textsuperscript{1}, Jiefu Li\textsuperscript{1}, Zhenfang Liu\textsuperscript{1}, \\ Jayden Chen\textsuperscript{3}, Zhen Bi\textsuperscript{1\textdagger}, Jungang Lou\textsuperscript{1}%
\thanks{* Equal contribution. \quad \textdagger\ Corresponding author.}
\end{tabular}}
\address{%
\textsuperscript{1}Huzhou Normal University
\textsuperscript{2}Alibaba Group
\textsuperscript{3}University of Waterloo}
\begin{document}
\ninept
\maketitle
\begin{abstract}
Long-horizon tasks require preserving and later recovering cross-session evidence under a bounded, query-blind memory budget. Existing compression can discard fine-grained visual cues or conflate semantically similar but incompatible observations. We present \textbf{C3M}, a cross-session multimodal memory organization that maintains a bounded active index over persistent source text-image evidence. Relation-aware updates consolidate safe redundancy while preserving complementary and incompatible records. At query time, budgeted routing selects useful index pages and expands their associated source evidence under a fixed reader budget. Together, these mechanisms establish a compact, provenance-preserving multimodal memory organization for cross-session long-horizon tasks, retaining temporal distinctions and source links required for reliable downstream reasoning. Code is available at \url{https://github.com/HuzhouNLP/C3M}.
\end{abstract}
\section{Introduction}
\label{sec:intro}

Long-horizon tasks require agents to preserve information across interactions \cite{dong2026longhorizon,park2023generative,chhikara2025mem0},
because later requests may depend on earlier observations, changed states, or
evidence whose relevance becomes apparent only retrospectively
\cite{wang2023longmem,gutierrez2024hipporag,xu2025amem}. 
This has motivated persistent memory systems that
store, update, organize, and retrieve accumulated experience across time
\cite{zhong2024memorybank,
tan2025rmm,ma2026nemori,wang2024memoryllm}. The challenge becomes harder in cross-session multimodal
settings, where relevant evidence may be distributed across text and images, and
fine-grained visual details can determine future answers. Memory must therefore
remain compact while keeping earlier source evidence recoverable when summaries
are insufficient.

\begin{figure}[!t]
    \centering
    \includegraphics[width=0.85\columnwidth]{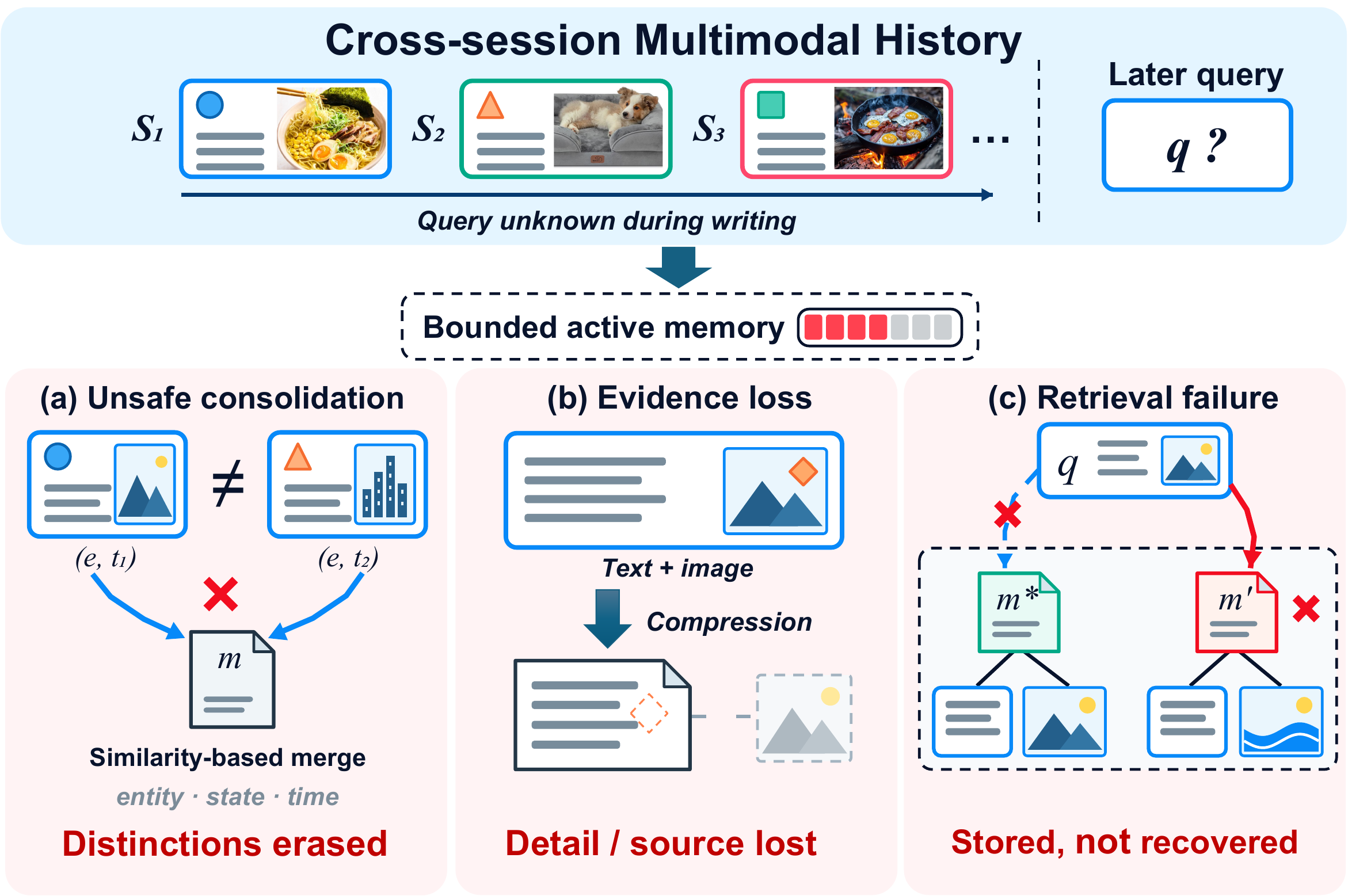}\par\vspace{-3pt}
    \caption{\textbf{Query-blind multimodal memory challenges.} (a) Similar
    records can be distinct. (b) Compression can lose visual details or source
    links. (c) Routing can leave stored evidence unrecovered.}
    \label{fig:intro_challenges}
\end{figure}

\begin{figure*}[!t]
    \centering
    \includegraphics[width=0.85\textwidth]{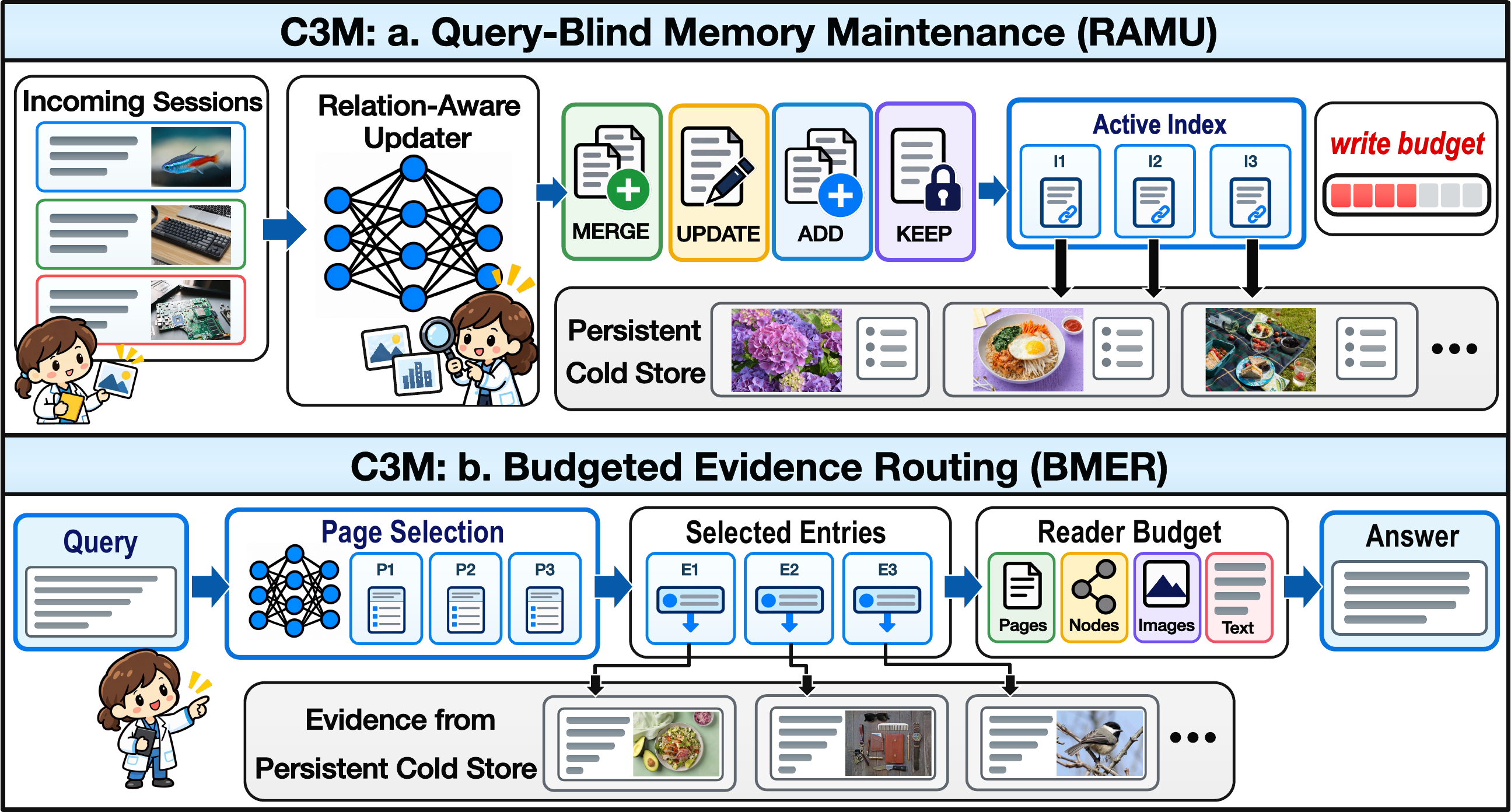}\par\vspace{-5pt}
    \caption{\textbf{Overview of C3M.} (a) During query-blind memory maintenance, RAMU applies relation-aware \textsc{Merge}, \textsc{Update}, \textsc{Add}, or \textsc{Keep} actions to incoming multimodal sessions. Compact, provenance-linked entries are maintained in the active index under a fixed write budget, while original evidence is preserved in the persistent Cold Store. (b) Given a query, BMER selects index pages (P) and entries (E), then retrieves linked Raw Nodes and original images from the Cold Store under a joint reader budget over pages, nodes, images, and text for answer generation.}
    \label{fig:main}
\end{figure*}



The central challenge is \textbf{maintaining a bounded, query-blind state without collapsing distinctions that future queries may require}. 
Observations can be redundant or complementary, while records similar in text or appearance may
differ in entity, state, or time.
Aggressive merging can erase differences visible
only in joint text-image context; retaining every near-duplicate instead consumes capacity and weakens routing selectivity. Compact entries must therefore preserve
provenance, distinguish stable facts from later states, and recover source images when summaries are insufficient. 
These requirements are especially important
across sessions: interruptions, partial observability, and delayed feedback make
forgotten or conflated state difficult to reconstruct, while irreversible actions
amplify the cost of errors \cite{dong2026longhorizon}.


Existing multimodal memory systems address parts of this problem through
visual-token compression and sparse memory, fixed-budget
updates, and gist or context distillation
\cite{song2024moviechat,song2026causalmem,lian2026mmmem,wu2025resum}. 
Content-aware and conflict-aware \cite{kim2026memrefine,ma2026conflict} approaches improve retention, but similarity alone does not imply safe merging. Compression may discard visual cues,
text-image bindings, source images, or temporal versions, while routing remains constrained by fixed reader budgets. 
Memory benefits also vary, reflecting trade-offs among compactness, information preservation, and utility \cite{bi2026conditional,wang2026contextcompression}. 
Multimodal long-context reasoning is further challenged by image-text distractors, compression-induced loss of visual fidelity, and modality-relevance gaps \cite{wu2025visualhaystacks,ren2026memlens,kang2026vmem}.

Therefore, we  introduce \textbf{C3M}, a cross-session
multimodal memory organization that maintains a bounded active index over persistent source text-image evidence, preserves necessary distinctions during
maintenance, and expands source evidence from selected pages under a fixed reader budget.
{Our detailed contributions are as follows:}

\begin{itemize}
    \setlength{\itemsep}{1pt}
    \setlength{\topsep}{1pt}
    \item \textbf{A multimodal memory organization for cross-session long-horizon tasks}
    that couples a bounded active index with persistently addressable text-image evidence.
    \item \textbf{C3M for multimodal memory maintenance.}
    It maintains compact routing representations over persistent source evidence accumulated across sessions.
    \item \textbf{More effective memory organization under a fixed memory budget}
    than existing compression methods, preserving evidence distinctions and
    provenance-linked access to original images from earlier sessions.
\end{itemize}

\newpage
\begin{table*}[!t]
    \centering
    \caption{Main results under our online, session-by-session evaluation protocol on MemLens, across two execution models, five task categories, and three history lengths. All reported scores are accuracy percentages, shown without the \% sign; boldface indicates the best scores among the compared methods.}
    \label{tab:main_results}
    \setlength{\tabcolsep}{1.5pt}
    \renewcommand{\arraystretch}{1.08}
    \footnotesize
    \resizebox{\textwidth}{!}{%
    \begin{tabular}{@{}ll*{18}{c}@{}}
        \toprule
        \multirow{2}{*}{\textbf{Model}} &
        \multirow{2}{*}{\textbf{Method}} &
        \multicolumn{3}{c}{\textbf{IE}} &
        \multicolumn{3}{c}{\textbf{MSR}} &
        \multicolumn{3}{c}{\textbf{TR}} &
        \multicolumn{3}{c}{\textbf{KU}} &
        \multicolumn{3}{c}{\textbf{AR}} &
        \multicolumn{3}{c}{\textbf{Overall}} \\
        \cmidrule(lr){3-5}\cmidrule(lr){6-8}\cmidrule(lr){9-11}\cmidrule(lr){12-14}\cmidrule(lr){15-17}\cmidrule(l){18-20}
        & & \textbf{32K} & \textbf{64K} & \textbf{128K} &
        \textbf{32K} & \textbf{64K} & \textbf{128K} &
        \textbf{32K} & \textbf{64K} & \textbf{128K} &
        \textbf{32K} & \textbf{64K} & \textbf{128K} &
        \textbf{32K} & \textbf{64K} & \textbf{128K} &
        \textbf{32K} & \textbf{64K} & \textbf{128K} \\
        \midrule
        \multirow{5}{*}{GPT-5.6 Sol} & ReSum & 44.26 & 37.70 & 42.62 & 22.86 & 20.00 & 22.86 & 64.58 & 56.25 & 43.75 & 34.48 & 24.14 & 17.24 & 59.09 & 50.00 & 45.45 & 45.64 & 38.46 & 35.90 \\
        & MovieChat & \textbf{73.77} & 65.57 & 57.38 & 45.71 & 42.86 & 54.29 & 70.83 & 70.83 & 58.33 & 44.83 & 48.28 & 44.83 & 86.36 & 86.36 & 72.73 & 65.13 & 62.56 & 56.92 \\
        & MemRefine & 67.21 & 60.66 & 65.57 & 65.71 & 57.14 & \textbf{57.14} & 72.92 & 75.00 & 70.83 & 51.72 & 51.72 & 44.83 & 86.36 & 81.82 & 63.64 & 68.21 & 64.62 & 62.05 \\
        & CDRs & 72.13 & 67.21 & \textbf{68.85} & 45.71 & 51.43 & 51.43 & 64.58 & 66.67 & \textbf{72.92} & 48.28 & 51.72 & \textbf{51.72} & 90.91 & 86.36 & 81.82 & 64.10 & 64.10 & \textbf{65.64} \\
        \rowcolor{gray!15}
        & \shortstack[l]{\textbf{C3M} \textbf{(Ours)}} & 72.13 & \textbf{68.85} & 60.66 & \textbf{71.43} & \textbf{57.14} & 54.29 & \textbf{77.08} & \textbf{79.17} & 70.83 & \textbf{55.17} & \textbf{55.17} & 48.28 & \textbf{95.45} & 77.27 & \textbf{86.36} & \textbf{73.33} & \textbf{68.21} & 63.08 \\
        \midrule
        \multirow{5}{*}{Qwen 3.8 Flash} & ReSum & 21.67 & 21.67 & 18.33 & 29.41 & 26.47 & 26.47 & 54.35 & 54.35 & 45.65 & 39.29 & 21.43 & 17.86 & 57.14 & \textbf{80.95} & \textbf{76.19} & 37.57 & 37.04 & 32.80 \\
        & MovieChat & 68.33 & 58.33 & 46.67 & 44.12 & 44.12 & 52.94 & 67.39 & 73.91 & 50.00 & 50.00 & \textbf{57.14} & 46.43 & 61.90 & 61.90 & 66.67 & 60.32 & 59.79 & 50.79 \\
        & MemRefine & 68.33 & 60.00 & \textbf{56.67} & 50.00 & \textbf{55.88} & \textbf{55.88} & 71.74 & 71.74 & 63.04 & 57.14 & 50.00 & 46.43 & 76.19 & 66.67 & 57.14 & 65.08 & 61.38 & 56.61 \\
        & CDRs & 68.33 & 65.00 & 55.00 & 44.12 & 50.00 & 52.94 & 71.74 & 69.57 & 69.57 & 57.14 & 50.00 & 46.43 & 66.67 & 71.43 & 57.14 & 62.96 & 61.90 & \textbf{57.14} \\
        \rowcolor{gray!15}
        & \shortstack[l]{\textbf{C3M} \textbf{(Ours)}} & 63.33 & \textbf{68.33} & 51.67 & \textbf{67.65} & 50.00 & 47.06 & \textbf{76.09} & \textbf{76.09} & \textbf{69.57} & 50.00 & 42.86 & \textbf{46.43} & \textbf{76.19} & 66.67 & 66.67 & \textbf{66.67} & \textbf{62.96} & 56.08 \\
        \bottomrule
    \end{tabular}
    }
\end{table*}

\section{Cross-Session Multimodal Memory Maintenance}
\label{sec:method}

Let $\mathcal{S}_{1:t}$ denote the sessions observed up to time $t$. Each session
yields immutable \emph{Raw Nodes} of text, images, and provenance in a persistent
Cold Store, whereas the bounded active index $\mathcal{I}_t$ stores only compact
routing representations and pointers. Built before the final query $q$ is known,
C3M uses \textbf{Relation-Aware Multimodal Memory Update (RAMU)} for each arriving session
and \textbf{Budgeted Multimodal Evidence Routing (BMER)} for $q$ (Fig.~\ref{fig:main}).
\textbf{BMER} traverses the index under a joint page, node, image, and text budget
$\mathcal{B}$: \textbf{RAMU} preserves write-time distinctions, and \textbf{BMER} retrieves their
original evidence when needed.

\subsection{Relation-Aware Multimodal Memory Update}
\label{sec:ramu}

When a new session $\mathcal{S}_t$ arrives, \textbf{RAMU} constructs an incoming entry
$m_t$ with routing representation $\mathbf{z}_t$. Let $\mathcal{U}_t$ denote its
Raw Nodes; the entry stores their pointers rather than replacing their contents:
\begin{equation}
\begin{aligned}
m_t &= (\mathbf{z}_t,\mathcal{P}_t,\tau_t), \\
\mathcal{P}_t &= \{\operatorname{ptr}(u)\mid u\in\mathcal{U}_t\},
\qquad \mathcal{U}_{1:t-1}\subseteq\mathcal{U}_{1:t}.
\end{aligned}
\label{eq:ramu-entry}
\end{equation}
It retrieves a small candidate set and predicts a relation and confidence for
every candidate as
\begin{equation}
\begin{aligned}
\mathcal{N}(m_t) &= \operatorname{TopK}_{e\in\mathcal{I}_{t-1}}
    s(\mathbf{z}_t,\mathbf{z}_e), \\
r_{t,e} &= \underset{r\in\mathcal{R}}{\operatorname*{arg\,max}}\;
    g_r\bigl(\phi(m_t,e)\bigr), \\
c_{t,e} &= \max_{r\in\mathcal{R}} g_r\bigl(\phi(m_t,e)\bigr).
\end{aligned}
\label{eq:ramu}
\end{equation}
Here, $s$ retrieves candidates; $\phi$ compares entity, state, time, action,
scope, and text-image agreement; and $\mathcal{R}$ contains same, update,
distinct, and uncertain.
Following the prospective local-update principle of reflective memory management
\cite{tan2025rmm}, similarity proposes candidates rather than deciding a merge.
For the highest-confidence candidate $e_t^\star$, \textbf{RAMU} takes a conservative
action under threshold $\delta$:
\begin{equation}
a_t =
\begin{cases}
\textsc{Merge}(e_t^\star), & r_{t,e_t^\star}=\mathrm{same}\ \land\ c_{t,e_t^\star}\geq\delta, \\
\textsc{Update}(e_t^\star,m_t), & r_{t,e_t^\star}=\mathrm{update}\ \land\ c_{t,e_t^\star}\geq\delta, \\
\textsc{Add}(m_t), & \mathcal{N}(m_t)=\emptyset, \\
\textsc{Keep}(m_t), & \text{otherwise}.
\end{cases}
\label{eq:ramu-action}
\end{equation}
A compatible description of the same stable fact is therefore \textsc{Merge}; a
later state of the same entity is \textsc{Update}; and related but
non-interchangeable or uncertain evidence is \textsc{Keep}. An update creates
links to both the predecessor and the new source, whereas a keep action retains an
independent entry. In all cases, Raw Nodes are never overwritten.
Let $\operatorname{Src}(e)$ denote the source pointers reachable from entry $e$,
including predecessor links. For a merged or updated entry $e'_t$, provenance
preservation requires
\begin{equation}
\operatorname{Src}(e'_t)\supseteq
\operatorname{Src}(e_t^\star)\cup\mathcal{P}_t.
\label{eq:source-preservation}
\end{equation}
This condition preserves access to both earlier and incoming evidence without
requiring identical routing representations. A merge consolidates support for a
stable fact, whereas an update retains the connection between successive states.
For \textsc{Add} and \textsc{Keep}, $\mathcal{P}_t$ remains attached to an independent
entry. Thus, maintenance reorganizes access while preserving the underlying
observations for queries not yet known.

At capacity, C3M consolidates only confirmed safe redundancy. Otherwise, it uses a
separately recorded pointer-preserving fallback rather than a semantic merge,
preventing surface similarity from erasing necessary distinctions.

\subsection{Budgeted Multimodal Evidence Routing}
\label{sec:bmer}

At query time, \textbf{BMER} routes $q$ through page, entry, Raw Node, and original
text-image addresses. Let $\mathcal{P}$ be the index pages and
$\operatorname{Ent}(p)$ the entries on page $p$. It first forms a hierarchy of
candidate pages, entries, and pointer-linked evidence groups:
\begin{equation}
\begin{aligned}
\mathcal{P}_q &= \operatorname{TopK}_{p\in\mathcal{P}}
\left[s(q,p)+\gamma\operatorname{CovGain}(p)\right], \\
\mathcal{D}_q &= \operatorname{TopK}_{e\in\bigcup_{p\in\mathcal{P}_q}\operatorname{Ent}(p)}
s(q,e), \\
\mathcal{C}_q &= \bigcup_{e\in\mathcal{D}_q}\operatorname{Expand}(e).
\end{aligned}
\label{eq:bmer-candidates}
\end{equation}
$\operatorname{CovGain}$ favours source sessions, recorded relation labels, and evidence
clusters not yet represented, while $\operatorname{Expand}$ resolves an entry into
its associated Raw Nodes, text, and original images. From $\mathcal{C}_q$, it
selects:
\begin{equation}
\begin{aligned}
\mathcal{E}_q^\star &=
\underset{\mathcal{E}\subseteq\mathcal{C}_q}{\operatorname*{arg\,max}}\;
\left[\operatorname{Rel}(q,\mathcal{E})+
\lambda\operatorname{Cov}(\mathcal{E})\right] \\
\text{s.t.}\quad \operatorname{Cost}(\mathcal{E}) &\preceq \mathcal{B}.
\end{aligned}
\label{eq:bmer}
\end{equation}
$\operatorname{Cov}$ rewards uncovered source sessions, recorded relation labels, and
evidence clusters, while $\lambda$ sets the relevance--coverage trade-off. More
explicitly, the joint read budget is
\begin{equation}
\begin{aligned}
\operatorname{Cost}(\mathcal{E}) =
\bigl(&|\operatorname{Pg}(\mathcal{E})|,
|\operatorname{Raw}(\mathcal{E})|, \\
&|\operatorname{Img}(\mathcal{E})|,
\sum_{u\in\operatorname{Raw}(\mathcal{E})}|\operatorname{Txt}(u)|\bigr)
\preceq \mathcal{B}.
\end{aligned}
\label{eq:bmer-cost}
\end{equation}
Entries only guide routing, while selected Raw Nodes and original images reach the
answerer. Every accessed page, node, image, and text token is charged to
$\mathcal{B}$. It logs page-entry retrieval,
Raw Node-image expansion, and final context coverage to separate update, routing,
and answerer failures.
The hierarchy separates locating evidence from reading it. For the selected
set $\mathcal{E}_q^\star$, define the source context and final answer as
\begin{equation}
\begin{aligned}
\mathcal{X}_q &= \operatorname{Text}(\mathcal{E}_q^\star)
\cup \operatorname{Images}(\mathcal{E}_q^\star), \\
\hat{y}_q &= \operatorname{Answer}(q,\mathcal{X}_q).
\end{aligned}
\label{eq:source-answer}
\end{equation}
Here, $\operatorname{Text}$ and $\operatorname{Images}$ return source text and
original images. Coverage preserves cross-session evidence, while
Eq.~\eqref{eq:bmer-cost} bounds the source context even after an entry is selected.

\begin{figure*}[!t]
\centering
\includegraphics[width=\textwidth]{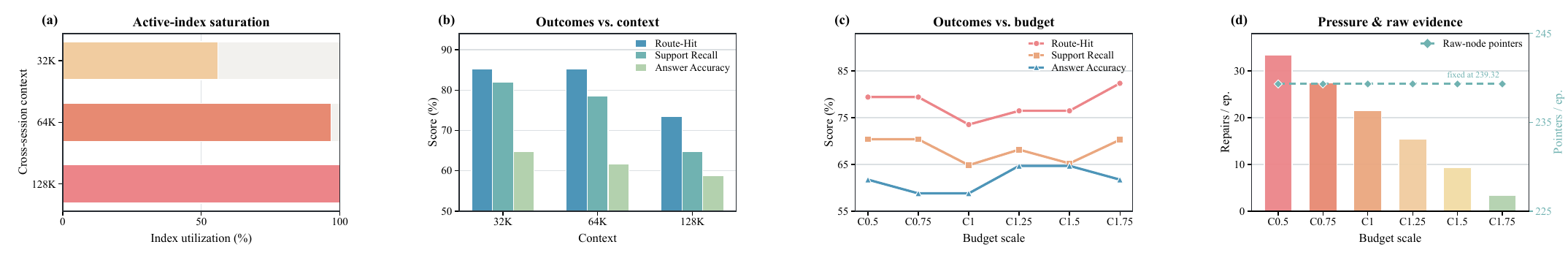}\par\vspace{-3pt}
\caption{Cross-session history pressure and active-index budget scaling. (a, b) With fixed C1 limits, history growth raises utilization and lowers task outcomes. (c, d) At fixed 128K history, active-index scaling reduces overflow repairs but yields non-monotonic outcomes; Raw-Node pointers remain fixed. Writer policy and reader budget are shared.}
\label{fig:ablation}
\vspace{-6pt}
\end{figure*}

\begin{figure*}[!t]
\centering
\includegraphics[width=\textwidth]{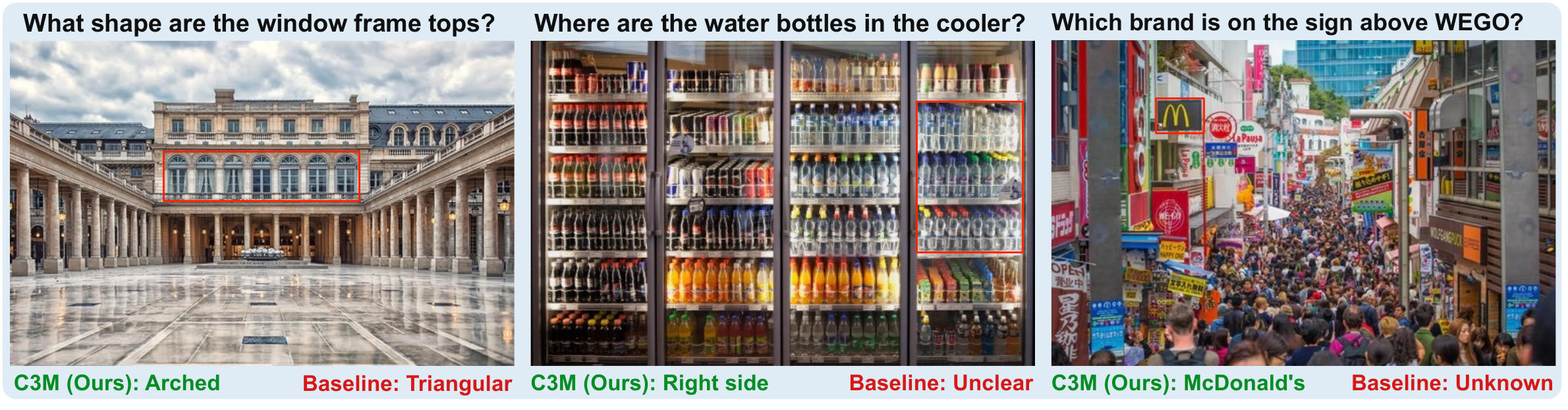}\par\vspace{-4pt}
\caption{Case studies of multimodal evidence routing. C3M correctly identifies arched window frames (left), water bottles on the cooler's right side (middle), and the McDonald's sign above WEGO (right), while the baseline gives incorrect or inconclusive answers. Red boxes mark relevant evidence; green and red text indicate C3M and baseline answers, respectively.}
\label{fig:case_study}
\vspace{-6pt}
\end{figure*}

\section{Experiment}
\subsection{Experimental Settings}
We evaluate MemLens's official 195-question agent subset (789 total)
at 32K/64K/128K histories \cite{ren2026memlens}.
Tasks cover information extraction (IE), multi-session reasoning (MSR),
temporal reasoning (TR), knowledge update (KU), and answer refusal (AR),
potentially combining visual details with textual context across sessions.

Unlike offline full-history evaluation, chronological replay retains original
questions and interleaved text-image evidence. Updates access only incoming
sessions and memory; future sessions, query, gold answer, and question type are
hidden. Query-blind consolidation and forgetting precede budgeted retrieval
after the final session.

\subsection{Baselines and Fair Protocol}
Baselines retain their policies over shared Raw Nodes and provenance:
text-based ReSum summarizes older blocks, retaining recent nodes
\cite{wu2025resum}; video-based MovieChat merges temporally adjacent entries
in short-/long-term buffers \cite{song2024moviechat}; text-based MemRefine applies
LLM-guided \textsc{Delete}/\textsc{Merge}/\textsc{Preserve} to globally similar pairs
\cite{kim2026memrefine}; embodied-memory CDRs uses an independent graph with
query-local conflict detection \cite{ma2026conflict}.

Shared limits cover the active store (36 entries, six
pages, 6,144 index tokens, 384~KiB), and reader limits (page beam 3, 12 Raw Nodes,
eight images, 12K text tokens). Answerers match per model; one judge is shared. CDRs's graph bytes and pre-filter candidates count toward the
same budgets. Raw Nodes and embeddings are shared; only maintenance and routing differ.
This controls evidence availability when comparing how methods organize and retrieve memory.

\subsection{Main Results}
In Table~\ref{tab:main_results}, GPT-5.6 Sol/Qwen 3.8 Flash build and maintain memory
and answer queries; one frontier model judges both.

C3M leads Overall at 32K and 64K with both models. With GPT-5.6 Sol, C3M leads or ties
on MSR/TR/KU at both lengths; with Qwen 3.8 Flash, it leads on MSR/TR/AR
at 32K and IE/TR at 64K, supporting temporal distinctions and cross-session
evidence recovery across both execution models. At 128K, CDRs leads Overall; C3M leads
GPT-5.6 Sol AR and ties on Qwen 3.8 Flash KU. Multimodal memory maintenance
leaves room for further optimization as cross-session histories grow longer.

\section{Ablation Study}
\subsection{Cross-Session Context Pressure}
\textbf{Longer cross-session histories increase active-index pressure.}
We conduct this diagnostic on a fixed, coverage-balanced subset of 34 questions
spanning all five MemLens abilities. All conditions use the same query-blind
writer, fixed C1 (24-entry) active-index budget, and reader allocation.
Fig.~\ref{fig:ablation} shows that, as history grows from 32K to 128K,
active-index utilization reaches saturation while Route-Hit, Support Recall,
and Answer Accuracy decline. The accompanying resource audit reports a
corresponding increase in overflow repairs, from 0.00 to 21.44 per episode,
while the reader allocation remains unchanged. Raw-Node pointers increase
with the source history (64.79, 126.97, and 239.32 per episode) and match the
number of input Raw Nodes, indicating that the active-index constraint limits
routing organization rather than deleting the underlying source text-image
evidence.

\subsection{Active-Index Budget Scaling}
\textbf{Active-index expansion reduces maintenance pressure but does not
guarantee monotonic task gains.}
At a fixed 128K cross-session history length, we proportionally scale the
entry capacity, per-page capacity, routing-text allowance, and index-byte
allowance, while holding the reader budget and persistent Raw-Node pointers
constant. Fig.~\ref{fig:ablation} shows that expanding the active index
steadily reduces overflow repairs, from 33.38 to 3.38 per episode. In
contrast, Route-Hit, Support Recall, and Answer Accuracy vary non-monotonically
and peak at different scales: Route-Hit is highest at C1.75, whereas Overall
accuracy peaks at C1.25 and C1.5. Thus, additional capacity alleviates
structural index pressure but cannot replace selective index organization and
query-time evidence routing.

\subsection{Case Study: Multimodal Evidence Routing}
\textbf{C3M routes queries to original multimodal evidence.}
In Fig.~\ref{fig:case_study}, C3M identifies arched window frames, water bottles
on the cooler's right side, and McDonald's above WEGO; the baseline answers
triangular, unclear, and unknown, respectively. Compact index entries and
provenance pointers guide budgeted retrieval of Raw Nodes and original images
from the multimodal memory store to verify shape, location, and brand against
source evidence.

\section{Conclusion}
We proposed a multimodal memory organization for cross-session long-horizon tasks.
It maintains memory under a fixed active-index budget without access to future
queries, coupling a bounded active index with persistently addressable source
text-image evidence:
relation-aware maintenance preserves distinctions among redundant, evolving, and
incompatible observations, while budgeted evidence routing connects compact index
entries to the original text and images through provenance-preserving pointers.
Experiments under our online, session-by-session evaluation protocol across
different cross-session histories show that C3M performs better overall than other
memory compression methods under shared index and reader budgets. C3M integrates
source evidence preservation, relation-aware updates, and budgeted retrieval
for cross-session multimodal memory maintenance in long-horizon tasks.

\clearpage
\section{Acknowledgment}
This work was supported by the National Natural Science Foundation of China under Grant 62506128, the Zhejiang Provincial Natural Science Foundation of China under Grants LQN25F020023 and LRG25F030003, and the Huzhou Key Research and Development Program under Grant 2025YZ23. The authors declare no competing interests.

\section{Compliance with Ethical Standards}
This study used publicly available data and did not involve the collection of new data from human participants or animals. Ethical approval was not required for this study.

\bibliographystyle{IEEEbib}
\bibliography{refs}

@inproceedings{chhikara2025mem0,
  author    = {Prateek Chhikara and
               Dev Khant and
               Saket Aryan and
               Taranjeet Singh and
               Deshraj Yadav},
  title     = {Mem0: Building Production-Ready {AI} Agents with Scalable Long-Term
               Memory},
  booktitle = {{ECAI}},
  series    = {Frontiers in Artificial Intelligence and Applications},
  volume    = {413},
  pages     = {2993--3000},
  publisher = {{IOS} Press},
  year      = {2025}
}

@inproceedings{xu2025amem,
  author    = {Wujiang Xu and
               Zujie Liang and
               Kai Mei and
               Hang Gao and
               Juntao Tan and
               Yongfeng Zhang},
  title     = {A-Mem: Agentic Memory for {LLM} Agents},
  booktitle = {NeurIPS},
  year      = {2025}
}

@inproceedings{tan2025rmm,
  author    = {Zhen Tan and
               Jun Yan and
               I{-}Hung Hsu and
               Rujun Han and
               Zifeng Wang and
               Long T. Le and
               Yiwen Song and
               Yanfei Chen and
               Hamid Palangi and
               George Lee and
               Anand Rajan Iyer and
               Tianlong Chen and
               Huan Liu and
               Chen{-}Yu Lee and
               Tomas Pfister},
  title     = {In Prospect and Retrospect: Reflective Memory Management for Long-term
               Personalized Dialogue Agents},
  booktitle = {{ACL} {(1)}},
  pages     = {8416--8439},
  publisher = {Association for Computational Linguistics},
  year      = {2025}
}

@inproceedings{ma2026nemori,
  author    = {Wenquan Ma and
               Jiayan Nan and
               Wenlong Wu},
  title     = {What Deserves Memory: Adaptive Memory Distillation for {LLM} Agents},
  booktitle = {{ACL} {(1)}},
  pages     = {34789--34812},
  publisher = {Association for Computational Linguistics},
  year      = {2026}
}

@inproceedings{ma2026conflict,
  author    = {Kexin Ma and
               Haotian Wang and
               Shenglin Chen and
               Yishuai Cai and
               Yu Huang and
               Ruochun Jin},
  title     = {Conflict-Aware Memory for Embodied Agents: Enhancing Vector Data Quality
               via Detection Rules},
  booktitle = {{ACL} {(1)}},
  pages     = {28328--28347},
  publisher = {Association for Computational Linguistics},
  year      = {2026}
}

@inproceedings{song2024moviechat,
  author    = {Enxin Song and
               Wenhao Chai and
               Guanhong Wang and
               Yucheng Zhang and
               Haoyang Zhou and
               Feiyang Wu and
               Haozhe Chi and
               Xun Guo and
               Tian Ye and
               Yanting Zhang and
               Yan Lu and
               Jenq{-}Neng Hwang and
               Gaoang Wang},
  title     = {MovieChat: From Dense Token to Sparse Memory for Long Video Understanding},
  booktitle = {{CVPR}},
  pages     = {18221--18232},
  publisher = {{IEEE}},
  year      = {2024}
}

@inproceedings{lian2026mmmem,
  author    = {Niu Lian and
               Yuting Wang and
               Hanshu Yao and
               Jinpeng Wang and
               Bin Chen and
               Yaowei Wang and
               Min Zhang and
               Shu{-}Tao Xia},
  title     = {From Verbatim to Gist: Distilling Pyramidal Multimodal Memory via
               Semantic Information Bottleneck for Long-Horizon Video Agents},
  booktitle = {{ACL} {(1)}},
  pages     = {11601--11617},
  publisher = {Association for Computational Linguistics},
  year      = {2026}
}

@article{song2026causalmem,
  author  = {Baiyang Song and
             Yuli Lin and
             Qiong Wu and
             Tao Chen and
             Jun Peng and
             Xiao Chen and
             Yiyi Zhou and
             Rongrong Ji},
  title   = {Towards a Dynamic and Fixed-budget Memory Bank for Efficient Streaming
             Video Understanding},
  journal = {CoRR},
  volume  = {abs/2606.25658},
  year    = {2026}
}

@inproceedings{wu2025visualhaystacks,
  author    = {Tsung{-}Han Wu and
               Giscard Biamby and
               Jerome Quenum and
               Ritwik Gupta and
               Joseph E. Gonzalez and
               Trevor Darrell and
               David M. Chan},
  title     = {Visual Haystacks: {A} Vision-Centric Needle-In-A-Haystack Benchmark},
  booktitle = {{ICLR}},
  publisher = {OpenReview.net},
  year      = {2025}
}

@article{ren2026memlens,
  author  = {Xiyu Ren and
             Zhaowei Wang and
             Yiming Du and
             Zhongwei Xie and
             Chi Liu and
             Xinlin Yang and
             Haoyue Feng and
             Wenjun Pan and
             Tianshi Zheng and
             Baixuan Xu and
             Zhengnan Li and
             Yangqiu Song and
             Ginny Y. Wong and
             Simon See},
  title   = {MemLens: Benchmarking Multimodal Long-Term Memory in Large Vision-Language
             Models},
  journal = {CoRR},
  volume  = {abs/2605.14906},
  year    = {2026}
}

@article{kang2026vmem,
  author  = {Dingyi Kang and
             Dongming Jiang and
             Yi Li and
             Guanpeng Li and
             Bingzhe Li},
  title   = {V-Mem: Modality-Routed Retrieval for Long-Term Multimodal Agentic
             Memory},
  journal = {CoRR},
  volume  = {abs/2608.01543},
  year    = {2026}
}

@article{wu2025resum,
  author  = {Xixi Wu and
             Kuan Li and
             Yida Zhao and
             Liwen Zhang and
             Litu Ou and
             Huifeng Yin and
             Zhongwang Zhang and
             Yong Jiang and
             Pengjun Xie and
             Fei Huang and
             Minhao Cheng and
             Shuai Wang and
             Hong Cheng and
             Jingren Zhou},
  title   = {ReSum: Unlocking Long-Horizon Search Intelligence via Context Summarization},
  journal = {CoRR},
  volume  = {abs/2509.13313},
  year    = {2025}
}

@article{kim2026memrefine,
  author  = {Minjae Kim and
             Jinheon Baek and
             Soyeong Jeong and
             Sung Ju Hwang},
  title   = {MemRefine: LLM-Guided Compression for Long-Term Agent Memory},
  journal = {CoRR},
  volume  = {abs/2606.13177},
  year    = {2026}
}

@article{dong2026longhorizon,
  author  = {Guanting Dong and Xiaoshuai Song and Yuyang Hu and Jiajie Jin and Chenghao Zhang and Yifei Chen and Xiaoxi Li and Huaying Yuan and Xinyu Yang and Tongyu Wen and Jiejun Tan and Hongjin Qian and Shijue Huang and Junting Lu and Zhenyu Li and Wanjun Zhong and Yutao Zhu and Tat-Seng Chua and Zhicheng Dou and Ji-Rong Wen},
  title   = {Towards Long-Horizon Agents: A Survey},
  journal = {Preprints},
  year    = {2026},
  month   = {July},
  doi     = {10.20944/preprints202607.1328.v1},
  url     = {https://www.preprints.org/manuscript/202607.1328}
}

@article{wang2026contextcompression,
  author  = {Yifei Wang and Ziteng Wang and Yuling Shi and Silin Chen and Xinrui Wang and Yueqi Wang and Beijun Shen and Linjing Li and Xiaodong Gu and Julian McAuley and Daniel Dajun Zeng},
  title   = {Context Compression for {LLM} Agents: A Survey of Methods, Failure Modes, and Evaluation},
  journal = {Preprints},
  year    = {2026},
  month   = {May},
  doi     = {10.20944/preprints202605.2065.v1},
  url     = {https://doi.org/10.20944/preprints202605.2065.v1}
}

@article{bi2026conditional,
  author  = {Zhen Bi and
             Xueshu Chen and
             Yan Wang and
             Zhizhi Peng and
             Haosen Hong and
             Zhen Wang and
             Zhixuan Chu and
             Bingyu Zhu and
             Jungang Lou},
  title   = {Memory Is Not Always Needed: Characterizing Conditional Memory in
             Scientific Reasoning},
  journal = {CoRR},
  volume  = {abs/2608.23982},
  year    = {2026}
}

@inproceedings{park2023generative,
  author    = {Joon Sung Park and
               Joseph C. O'Brien and
               Carrie Jun Cai and
               Meredith Ringel Morris and
               Percy Liang and
               Michael S. Bernstein},
  title     = {Generative Agents: Interactive Simulacra of Human Behavior},
  booktitle = {{UIST}},
  pages     = {2:1--2:22},
  publisher = {{ACM}},
  year      = {2023}
}

@inproceedings{zhong2024memorybank,
  author    = {Wanjun Zhong and
               Lianghong Guo and
               Qiqi Gao and
               He Ye and
               Yanlin Wang},
  title     = {MemoryBank: Enhancing Large Language Models with Long-Term Memory},
  booktitle = {{AAAI}},
  pages     = {19724--19731},
  publisher = {{AAAI} Press},
  year      = {2024}
}

@inproceedings{wang2023longmem,
  author    = {Weizhi Wang and Li Dong and Hao Cheng and Xiaodong Liu and
               Xifeng Yan and Jianfeng Gao and Furu Wei},
  title     = {Augmenting Language Models with Long-Term Memory},
  booktitle = {NeurIPS},
  year      = {2023}
}

@inproceedings{wang2024memoryllm,
  author    = {Yu Wang and Yifan Gao and Xiusi Chen and Haoming Jiang and
               Shiyang Li and Jingfeng Yang and Qingyu Yin and Zheng Li and
               Xian Li and Bing Yin and Jingbo Shang and Julian McAuley},
  title     = {{MEMORYLLM}: Towards Self-Updatable Large Language Models},
  booktitle = {{ICML}},
  series    = {Proceedings of Machine Learning Research},
  volume    = {235},
  pages     = {50453--50466},
  publisher = {{PMLR}},
  year      = {2024}
}

@inproceedings{gutierrez2024hipporag,
  author    = {Bernal Jim{\'e}nez Guti{\'e}rrez and Yiheng Shu and Yu Gu and
               Michihiro Yasunaga and Yu Su},
  title     = {{HippoRAG}: Neurobiologically Inspired Long-Term Memory for
               Large Language Models},
  booktitle = {NeurIPS},
  year      = {2024}
}

\end{document}